\documentclass[runningheads,a4paper]{llncs}

\usepackage{amssymb}
\usepackage{graphicx}

\usepackage{url}
\urldef{\mailsa}\path|| \urldef{\mailsb}\path||
\urldef{\mailsc}\path||    
\newcommand{\keywords}[1]{\par\addvspace\baselineskip
\noindent\keywordname\enspace\ignorespaces#1}

\begin{document}

\mainmatter  % start of an individual contribution

\title{Audio Visual Speech Recognition using Deep Recurrent Neural Networks}

\titlerunning{Audio Visual Speech Recognition using Deep Recurrent Neural Networks}

\author{Abhinav Thanda \\
Shankar M Venkatesan}
\authorrunning{Audio Visual Speech Recognition using Deep Recurrent Neural Networks}

\institute{Samsung R\&D Institute India, Bangalore}

\toctitle{Lecture Notes in Computer Science}
\tocauthor{Authors' Instructions}
\maketitle
\pagenumbering{gobble}

\begin{abstract}
In this work, we propose a training algorithm for an audio-visual automatic speech recognition (AV-ASR) system using deep recurrent neural network (RNN).First, we train a deep RNN acoustic model with a Connectionist Temporal Classification (CTC) objective function. The frame labels obtained from the acoustic model are then used to perform a non-linear dimensionality reduction of the visual features using a deep bottleneck network. Audio and visual features are fused and used to train a fusion RNN. The use of bottleneck features for visual modality helps the model to converge properly during training. Our system is evaluated on GRID corpus. Our results show that presence of visual modality gives significant improvement in character error rate (CER) at various levels of noise even when the model is trained without noisy data. We also provide a comparison of two fusion methods: feature fusion and decision fusion.\footnote[1]{Version (Aug 2016) accepted in 4th International Workshop on Multimodal pattern recognition of social signals in human computer interaction(MPRSS 2016), a satellite event of the International Conference on Pattern Recognition (ICPR 2016)}

\end{abstract}

\section{Introduction}

Audio-visual automatic speech recognition (AV-ASR) is a case of multi-modal analysis in which two modalities (audio and visual) complement each other to recognize speech. Incorporating visual features, such as speaker's lip movements and facial expressions, into automatic speech recognition (ASR) systems has been shown to improve their performances especially under noisy conditions. To this end several methods have been proposed which traditionally include variants of GMM/HMM models\cite{dupont2000audio}\cite{brand1997coupled}. More recently AV-ASR methods based on deep neural networks (DNN) models\cite{huang2013audio}\cite{mroueh2015deep}\cite{noda2015audio} have been proposed. 

End-to-end speech recognition methods based on RNNs trained with CTC objective function\cite{graves2014towards}\cite{miao2015eesen}\cite{hannun2014deep} have come to the fore recently and have been shown to give performances comparable to that of DNN/HMM.  The RNN trained with CTC directly learns a mapping between audio feature frames and character/phoneme sequences. This method eliminates the need for intermediate GMM/HMM training thereby simplifying the training procedure. To our knowledge, so far AV-ASR systems based on RNN trained with CTC have not been explored.

\begin{figure}
	\centering
	\includegraphics[width=1.0\linewidth]{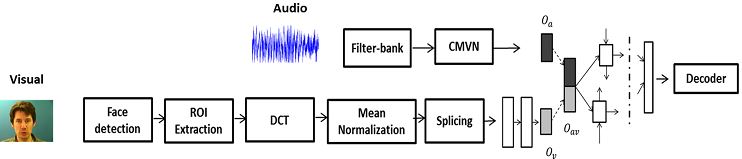}
	\caption{Pipeline of AV-ASR system using feature fusion method}
	\label{fig:avasr}
\end{figure}

In this work, we design and evaluate an audio-visual ASR (AV-ASR) system using deep recurrent neural network (RNN) and CTC objective function.  The design of an AV-ASR system includes the tasks of visual feature engineering, and audio-visual information fusion. Figure \ref{fig:avasr} shows the AV-ASR pipeline at test time. This work mainly deals with the visual feature extraction and processing steps and training protocol for the fusion model. Proper visual features are important especially in the case of RNNs as RNNs are difficult to train. Bottleneck features used in tandem with audio features are known to improve ASR performance \cite{gehring2013extracting}\cite{hermansky2000tandem}\cite{yu2011improved}. We employ a similar idea to improve the discriminatory power of video features. We show that this helps the RNN to converge properly when compared with raw features. Finally, we compare the performances of feature fusion and decision fusion methods.

The paper is organized as follows:  Section \ref{sssec:num2} presents the prior work on AV-ASR. Bi-directional RNN and its training using CTC objective function are discussed in Section \ref{sssec:num3}. Section \ref{sssec:num4} describes the feature extraction steps for audio and visual modalities. In section \ref{sssec:num5}  different fusion models are explained. Section \ref{sssec:num6} explains the training protocols and experimental results. Finally, we summarize our work in  \ref{sssec:num7}.

\section{Related Work} \label{sssec:num2}

The differences between various AV-ASR systems lie chiefly in the methods employed for visual feature extraction and audio-visual information fusion. Visual feature extraction methods can be of 3 types\cite{potamianos2003recent} : 1. Appearance based features where each pixel in the mouth region of the speaker (ROI) is considered to be informative. Usually a transformation such as DCT or PCA is applied to the ROI to reduce the dimensions. Additional feature processing such as mean normalization, intra-frame and inter-frame LDA may be applied \cite{huang2003improving}\cite{potamianos2003recent}.  2. Shape based features utilize the geometric features such as height, width and area of the lip region or build a statistical model of the lip contours whose parameters are used as features. 3. Combination of appearance and shape based features.

Fusion methods can be broadly divided into two types\cite{potamianos2003recent}\cite{katsaggelos2015audiovisual}: 1. Feature fusion 2. Decision fusion. Feature fusion models perform a low level integration of audio and visual features and this involves a single model which is trained on the fused features. Feature fusion may include a simple concatenation of features or feature weighting and is usually followed by a dimensionality reduction transformation like LDA. Decision fusion is applied in cases where the output classes for the two modalities are same. Various decision fusion methods based on variants of HMMs have been proposed\cite{dupont2000audio}\cite{brand1997coupled}. In Multistream HMM the emission probability of a state of audio-visual system is obtained by a linear combination of log-likelihoods of individual streams for that state. The parameters of HMMs for individual streams can be estimated separately or jointly. While multistream HMM assumes state level synchrony between the two streams, some methods\cite{bengio2004multimodal}\cite{brand1997coupled} such as coupled HMM\cite{brand1997coupled} allow for asynchrony between two streams. For a detailed survey on HMM based AV-ASR systems we refer the readers to  \cite{potamianos2003recent}\cite{katsaggelos2015audiovisual}

Application of deep learning to multi-modal analyses was presented in \cite{ngiam2011multimodal} which describes multi-modal, cross-modal and shared representation learning and their applications to AV-ASR. In \cite{huang2013audio},  Deep Belief Networks(DBN) are explored. In \cite{mroueh2015deep} the authors train separate networks for audio and visual inputs and fuse the final layers of two networks, and then build a third DNN with the fused features.  In addition, \cite{mroueh2015deep} presents a new DNN architecture with a bilinear soft-max layer which further improves the performance. In \cite{noda2015audio} a deep de-noising auto-encoder is used to learn noise robust speech features. The auto-encoder is trained with MFCC features of noisy speech as input and reconstructs clean features. The outputs of final layer of the auto-encoder are used as audio features. A CNN is trained with images from the mouth region as input and phoneme labels as output. The final layers of the two networks are then combined to train a multi-stream HMM.

\section{Sequence Labeling Using RNN}\label{sssec:num3}
The following notations are adopted in this paper. For an utterance $u$ of length $T_{u}$,  $\textbf{O}_{a}^{u}=(\overline{O}_{a,1}^{u}, \overline{O}_{a,2}^{u}, ... , \overline{O}_{a,T_{u}}^{u})$ and $\textbf{O}_{v}^{u}=(\overline{O}_{v,1}^{u}, \overline{O}_{v,2}^{u}, ... , \overline{O}_{v,T_{u}}^{u})$ denote the observation sequences of audio and visual frames where $\overline{O}_{a,t} \in \mathbb{R}^{d_{a}}$ and $\overline{O}_{v,t} \in \mathbb{R}^{d_{v}} $. We assume equal frame rates for audio and visual inputs which is ensured in experiments by means of interpolation. $\textbf{O}_{av}^{u}=(\overline{O}_{av,1}^{u}, \overline{O}_{av,2}^{u}, ... , \overline{O}_{av,T_{u}}^{u})$ where $\overline{O}_{av,t}^{u} = [\overline{O}_{a,t}^{u}, \overline{O}_{v,t}^{u}] \in \mathbb{R}^{d_{av}}$ where $d_{av}=d_{a}+d_{v}$ denotes the concatenated features at time t for utterance u. The corresponding label sequence is given by $l=(l_{1}, l_{2}, ..., l_{S_{u}})$ where $S_{u}\le T_{u}$ and $l_{i}\in L$ where $L$ is the set of English letters and an additional element representing a space. For ease of representation, we drop the utterance index $u$. All the models described in this paper are character based.
\subsection{Bi-directional RNN}
RNNs are a class of neural networks used to map sequences to sequences. This is possible because of the feedback connections between hidden nodes. In a bi-directional RNN, the hidden layer has two components each corresponding to forward(past) and backward(future) connections. For a given input sequence $\textbf{O}=(\overline{O}_{1}, \overline{O}_{2}, ... , \overline{O}_{T})$, the output of the network is calculated as follows:
forward pass through forward component of the hidden layer at a given instant $t$ is given by
\begin{equation}
\label{eq:eq8}
\overline{h}_{t}^{f}  = g(\textbf{W}_{ho}^{f}\overline{O}_{t}+\textbf{W}_{hh}^{f}\overline{h}_{t-1}^{f}+\overline{b}_{h}^{f})
\end{equation}
where $\textbf{W}_{ho}^{f}$ is the input-to-hidden weights for forward component, $\textbf{W}_{hh}^{f}$  corresponds to hidden-to-hidden weights between forward components, and $\overline{b}_{h}^{f}$ is the forward component bias. $g$ is a non-linearity depending on the choice of the hidden layer unit. Similarly, forward pass through the backward component of the hidden layer is given by
\begin{equation}
\label{eq:eq9}
\overline{h}_{t}^{b}  = g(\textbf{W}_{ho}^{b}\overline{O}_{t}+\textbf{W}_{hh}^{b}\overline{h}_{t-1}^{b}+\overline{b}_{h}^{b})
\end{equation}
where $\textbf{W}_{ho}^{b}$, $\textbf{W}_{hh}^{b}$, $\overline{b}_{h}^{b}$ are the corresponding parameters for the backward component. The input to next  layer is the concatenated vector $[\textbf{h}_{t}^{f},\textbf{h}_{t}^{b}]$. In a deep RNN multiple such bidirectional hidden layers are stacked. 

RNNs are trained using Back-Propagation Through Time (BPTT) algorithm. The training algorithm suffers from vanishing gradients problem which is overcome by using a special unit in hidden layer called the Long Short Term Memory(LSTM)\cite{hochreiter1997long}\cite{graves2012neural}.

\subsection{Connectionist Temporal Classification}
DNNs used in ASR systems are frame-level classifiers i.e., each frame of the input sequence is requires a class label in order for the DNN to be trained. The frame-level labels are usually HMM states, obtained by first training a GMM/HMM model and then by forced alignment of input sequences to the HMM states. CTC objective function\cite{graves2006connectionist}\cite{graves2014towards} obviates the need for such alignments as it enables the network to learn over all possible alignments.

Let the input sequence be $\textbf{O}=(\overline{O}_{1}, \overline{O}_{2}, ... , \overline{O}_{T})$ and a corresponding label sequence $\textbf{l}=(l_{1}, l_{2}, ..., l_{S})$ where $S\le T$. The RNN employs a soft-max output layer containing one node for each element in $L'$ where $L' = L \cup \{\phi\}$. The number of output units is $|L'|=|L|+1$. The additional symbol  $\phi$ represents a blank label meaning that the network has not produced an output for that input frame.  The additional blank label at the output allows us to define an alignment $\pi$ of length T containing elements of $L'$. For example, $(A \phi \phi M \phi), (\phi A \phi \phi M)$ are both alignments of length 5 for the label sequence $AM$.  Accordingly, a many to one map $B: L'^{T}  \longmapsto  L^{\le T}$ can be defined which generates the label sequence from an alignment.

Assuming that the posterior probabilities obtained at soft-max layer, at each instant are independent we get

\begin{equation} \label{eq:eq1} P(\pi | \textbf{O}) = \prod_{t=1}^{T} P(k_{t}| \overline{O}_{t}) \end{equation} where $k \in L'$ and 

\begin{equation} \label{eq:eq2} P(k_{t} | \overline{O}_{t}) = \frac{\exp(y_{t}^{k})}{\Sigma_{k'}\exp(y_{t}^{k'})} \end{equation}
where $y_{t}^{k}$ is the input to node $k$ of the
soft-max layer at time $t$

The likelihood of the label sequence given an observation sequence can be calculated by summing  (\ref{eq:eq1}) over all possible alignments.
\begin{equation} \label{eq:eq3} P(\textbf{l}|\textbf{O}) = \sum_{\pi \in B^{-1}(\textbf{l})} P(\pi | \textbf{O})  \end{equation}	

The goal is to maximize the log-likelihood $\log P(\textbf{l}|\textbf{O})$ estimation of a label sequence given an observation sequence. Equation \ref{eq:eq3} is computationally intractable since the number of alignments increases exponentially with the number of labels. 
For efficient computation of (\ref{eq:eq3}), forward-backward algorithm is used.

\section{Feature Extraction}\label{sssec:num4}

\subsection{Audio Features}
The sampling rate of audio data is converted to 16kHz. For each frame of speech signal  of 25ms duration, filter-bank features of 40 dimensions are extracted. The filter-bank features are mean normalized and $\Delta$ and $\Delta \Delta$ features are appended. The final 120 dimensional features are used as audio features.

\subsection{Visual Features}
The video frame rate is increased to match the rate of audio frames through interpolation. For AV-ASR, the ROI for visual features is the region surrounding the speaker's mouth. Each frame is converted to gray scale and face detection is performed using Viola-Jones algorithm. The 64x64 lip region is extracted by detecting 68 landmark points\cite{kazemi2014one} on the speakers face, and cropping the ROI surrounding speakers mouth and chin. 100 dimensional DCT features are extracted from the ROI. 

After several experiments of training with DCT features, we found that RNN training either exploded or converged poorly. In order improve the discriminatory power of the visual features, we perform non-linear dimensionality reduction of the features using a deep bottleneck network. Bottleneck features are obtained by training a neural network in which one of the hidden layers has relatively small dimensions. The DNN is trained using cross-entropy cost function with character labels as output. The frame-level character labels required for DNN training are obtained by first training an acoustic model ($RNN_{a}$) and then obtaining the outputs from the final soft-max layer of $RNN_{a}$. 

The DNN configuration is given by $dim-1024-1024-40-1024-opdim$ where $dim=1100$ and is obtained by splicing each 100 dimensional video frame with a context of 10 frames - 5 on each side. $opdim=|L'|$. After training, the last 2 layers are discarded and 40-dimensional outputs are used as visual features. The final dimension of visual feature vector is 120 including the $\Delta$ and $\Delta \Delta$ features.

\section{Fusion models}\label{sssec:num5}
\begin{figure}[t]
	\centering
	\includegraphics[width=0.7\linewidth]{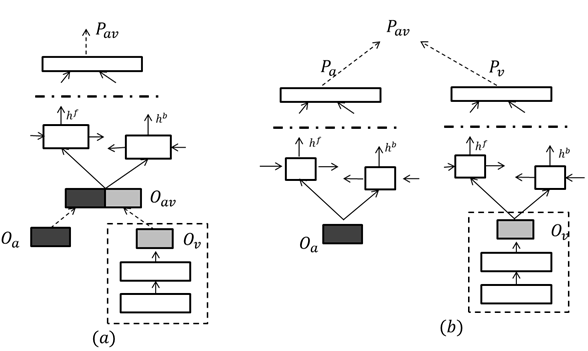}
	\caption{Fusion models (a) Featue fusion (b) Decision fusion. The bottleneck network for visual feature extraction is enclosed in the dotted box.}
	\label{fig:fusion}
\end{figure}

In this work, the fusion models are character based RNNs trained using CTC objective function i.e. $L'$ is the set of English alphabet including a blank label. The two fusion models are shown in Figure \ref{fig:fusion}

\subsection{Feature Fusion}
In feature fusion technique, a single $RNN_{av}$ is trained by concatenating the audio and visual features using the CTC objective function. In the test phase, at each instant the concatenated features are forward propagated through the network. In the CTC decoding step, the posterior probabilities obtained at the soft-max layer are converted to pseudo log-likelihoods\cite{vesely2013sequence} as 
\begin{equation}\label{eq:eq4} \log P_{av}(\overline{O}_{av,t} | k) = \log P_{av}(k | \overline{O}_{av,t})  - \log P(k)\end{equation} 

where $ k \in L' $ and $P(k)$ is the prior probability of class $k$ obtained from the training data \cite{miao2015eesen}.

\subsection{Decision Fusion}
In decision fusion technique the audio and visual modalities are modeled by separate networks, $RNN_{a}$ and $RNN_{v}$ respectively. 
$RNN_{v}$ is a lip-reading system. The networks are trained separately. In the test phase, for a given utterance the frame level, 
the pseudo log-likelihoods of $RNN_{a}$ and $RNN_{v}$ are combined as

\begin{equation}\label{eq:eq5}\log P_{av}(\overline{O}_{a,t}, \overline{O}_{v,t} | k) = \gamma \log P_{a}(k | \overline{O}_{a,t}) + (1 - \gamma) \log P_{v}(k | \overline{O}_{v,t})  - \log P(k)\end{equation}

where $0 \le \gamma \le 1$ is a parameter dependent on the noise level and the reliability of each modality\cite{dupont2000audio}. For example, at higher levels of noise in audio input, a low value of $\gamma$ is preferred. In this work, we adapt the parameter $\gamma$ for each utterance based on KL-divergence measure between the posterior probability distributions of $RNN_{a}$ and $RNN_{v}$. The divergence between the posterior probability distributions is expected to vary as the noise in the audio modality increases. The KL-divergence is scaled to a value in $[0,1]$ using logistic sigmoid. The parameter $b$ was determined empirically from validation dataset.
\begin{equation}\label{eq:eq6} D_{KL}(P_{v}||P_{a}) = \sum_{i} P_{v}log P_{a}\end{equation}
where we consider the posteriors of $RNN_{v}$ as the true distribution based on the assumption that video input is always free from noise.
\begin{equation}\label{eq:eq7} \gamma= \frac{1}{1+exp(-D_{KL}+b)}\end{equation}

\section{Experiments}\label{sssec:num6}
The system was trained and tested on GRID audio-visual corpus\cite{cooke2006audio}. GRID corpus is a collection of audio and video recordings of 34 speakers (18 male, 16 female) each uttering a 1000 sentences. Each utterance has a fixed length of approximately 3 seconds. The total number of words in the vocabulary is 51.  The syntactic structures of all sentences are similar as shown below. \newline \newline
$< command >$ \space $<color>$ \space $<preposition>$ \space $<letter>$ \space $<digit>$ \space $<adverb>$ \newline
Ex. PLACE RED AT M ZERO PLEASE\newline

\subsection{Training}
In the corpus obtained, the video recordings for speaker 21 were not available. In addition, 308 utterances by various speakers could not be processed due to various errors. The dataset in effect consisted of 32692 utterances 90\% of the which (containing 29423 utterances) was used for training and cross validation while the remaining (10\%) data was used as test set. Both training and test data contain utterances from all of the speakers.  Models were trained and tested using Kaldi speech recognition tool kit\cite{povey2011kaldi}, Kaldi+PDNN\cite{miao2014kaldi+} and EESEN framework\cite{miao2015eesen}. 
\subsubsection{$RNN_{a}$-Acoustic model}
$RNN_{a}$ contains 2 bi-directional LSTM  hidden layers.Input to the network is 120-dimensional vector containing filter-bank coefficients along with $\Delta$ and $\Delta \Delta$ features. The model parameters are randomly initialized within the range [-0.1,0.1]. The initial learning rate is set to 0.00004. Learning rate adaption is performed as follows: when the improvement in accuracy on the cross-validation set between two successive epochs falls below 0.5\%, the learning rate is halved.The halving continues for each subsequent epoch until the training stops when the increase in frame level accuracy is less than 0.1\%.
\subsubsection{Deep Bottleneck Network}
The training protocol similar to \cite{vesely2013sequence} was followed to train the bottleneck network. Input video features are mean normalized and spliced. Cross-entropy loss function is minimized using mini-batch Stochastic Gradient Descent (SGD). The frames are shuffled randomly before each epoch. Batch size is set to 256 and initial learning rate is set to 0.008. Learning rate adaptation similar to acoustic model is employed.
\subsubsection{$RNN_{v}$-Lip Reader}
\begin{figure}[t]
	\centering
	\includegraphics[width=0.7\linewidth]{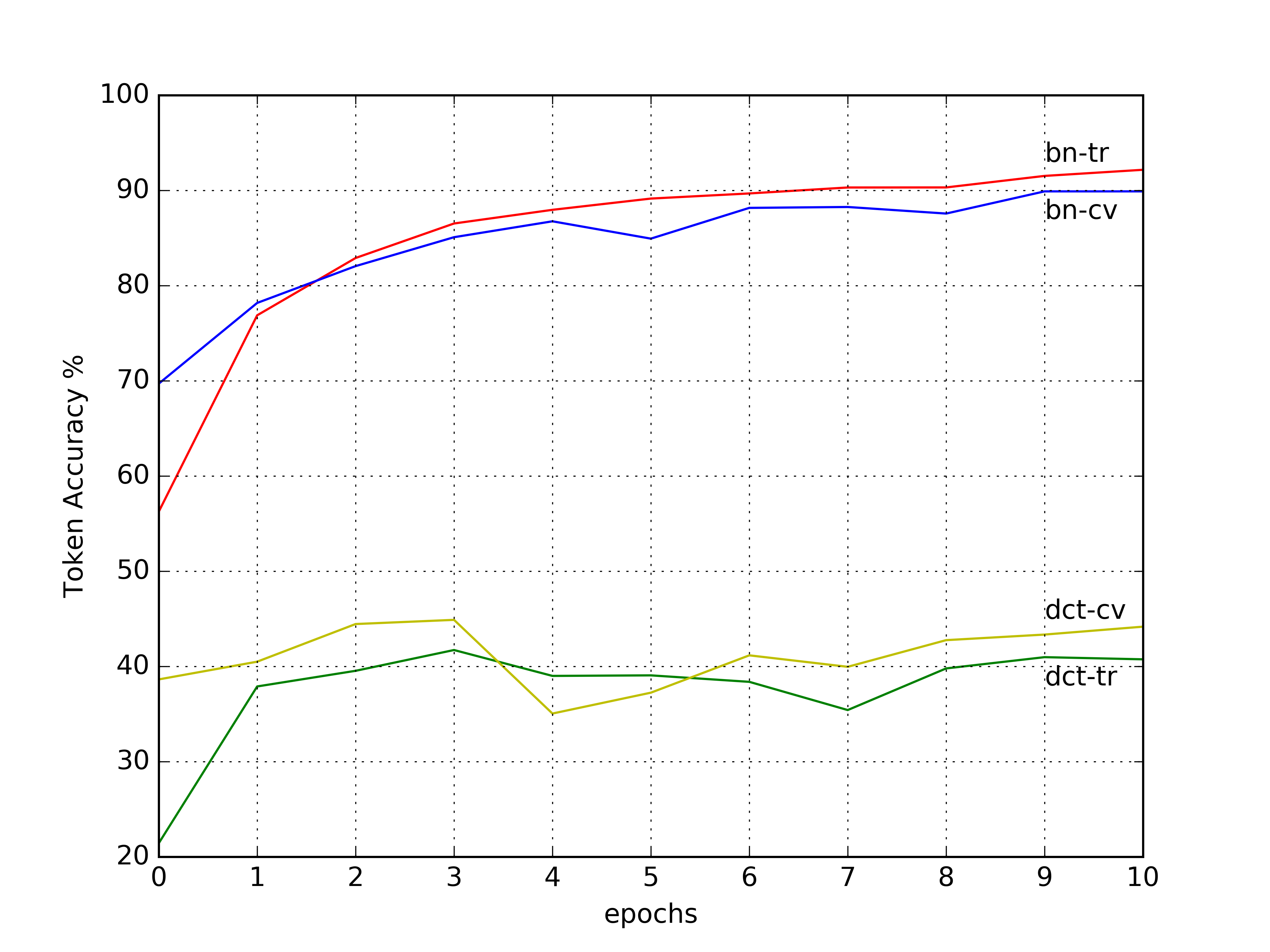}
	\caption{Learning curves for bottleneck(bn) features and DCT features for training(tr) and validation(cv) data sets.}
	\label{fig:dctvsbn}
\end{figure}

$RNN_{v}$ is trained with bottleneck network features as input. The network architecture and training procedure is same as $RNN_{a}$. Figure \ref{fig:dctvsbn} depicts the  learning curves when trained with bottleneck features and DCT features. The figure shows that bottleneck features are helpful in proper convergence of the model.

\subsubsection{$RNN_{av}$}
The feature fusion model $RNN_{av}$ consists of 3 bi-directional LSTM hidden layers. The input dimension is 240, corresponding to filter-bank coefficients of audio modality, bottleneck features of visual modality and their respective $\Delta$ features. The initialization and learning rate adaption are similar to acoustic model training. However, the learning rate adaptation is employed only after a minimum number of(in this case 20) epochs are completed. 

During each utterance in an epoch we first present the fused audio-visual fused input sequence followed by the input sequence with audio input set to very low values. This prevents the $RNN_{av}$ from over-fitting to audio only inputs. Thus the effective number of sequences presented to the network in a given epoch is twice the total number of training utterances (AV and V features). After the training with AV and V features we train the network once again with two epochs of audio only utterances obtained by turning off the visual modality. 

\subsection{Results}
 The audio-visual model is tested with three levels of babble noise 0dB SNR, 10dB SNR and clean audio. Noise was added to test data artificially by mixing babble noise with clean audio .wav files. In order to show the importance of visual modality under noisy environment, the model is tested with either audio or video inputs turned off.  A token WFST\cite{miao2015eesen} is used to map the paths to their corresponding label sequences. The token WFST obtains this mapping by removing all the blanks and repeated labels. Character Error Rate(CER) is obtained from the decoded and expected label sequences by calculating the Edit distance between them. The CER results are shown in Table \ref{table:FUSIONCOMPARE}.\begin{center}
 	\begin{table}
 		\centering
 		\caption[short text]{\% CER comparison for decision fusion($RNN_{a},RNN_{v}$) and feature fusion ($RNN_{av}$) models. $RNN_{a}$ 
 			is the acoustic model and $RNN_{v}$ is the lip reader}
 		\begin{tabular}{| l | l | l | l |}
 			\hline
 			Model & \multicolumn{2}{ c| }{Input} & CER \% \\ \hline
 			& Audio & Visual &  \\ \hline
 			$RNN_{av}$ & Clean & OFF & 7.35 \\ \hline 
 			$RNN_{av}$ & Clean & ON & 5.74 \\ \hline
 			$RNN_{av}$ & OFF & ON & 11.42 \\ \hline
 			$RNN_{av}$ & 10 SNR dB & OFF & 38.31 \\ \hline
 			$RNN_{av}$ & 10 SNR dB & ON & 10.24 \\ \hline
 			$RNN_{av}$ & 0 SNR dB & OFF & 59.65 \\ \hline
 			$RNN_{av}$ & 0 SNR dB & ON & 11.57 \\ \hline
 			$RNN_{a},RNN_{v}$ & Clean & OFF & 2.45 \\ \hline
 			$RNN_{a},RNN_{v}$ & Clean & ON & 8.46 \\ \hline
 			$RNN_{a},RNN_{v}$ & OFF & ON & 11.06 \\ \hline
 			$RNN_{a},RNN_{v}$ & 10 SNR dB & OFF & 23.83 \\ \hline
 			$RNN_{a},RNN_{v}$ & 10 SNR dB & ON & 14.83 \\ \hline
 			$RNN_{a},RNN_{v}$ & 0 SNR dB & OFF & 59.27 \\ \hline
 			$RNN_{a},RNN_{v}$ & 0 SNR dB & ON &  16.84 \\ \hline
 		\end{tabular}
 		\label{table:FUSIONCOMPARE}
 	\end{table}
 \end{center}
We observe that with clean audio input, audio only $RNN_{a}$ performs significantly better (CER 2.45\%) compared to audio-visual $RNN_{av}$ (CER 5.74\%). However as audio becomes noisy, the performance of $RNN_{a}$ deteriorates significantly whereas the performance  of $RNN_{av}$ remains relatively stable. Under noisy conditions the feature fusion model behaves as if it is not receiving any input from the audio modality. 

Table \ref{table:FUSIONCOMPARE} also gives a comparison between feature fusion model and decision fusion model. We find that feature fusion model performs better than decision fusion model in all cases except under clean audio conditions. The poor CER of $RNN_{a},RNN_{v}$ model indicates that the frame level predictions between $RNN_{a}$ and $RNN_{v}$ are not synchronous. However, both the fusion models provide significant gains under noisy audio inputs. While there is large difference between $RNN_{a}$ and other models with clean inputs, we believe this difference is due to the nature of dataset and will reduce with larger datasets. 

\section{Conclusions And Future Work}\label{sssec:num7}
In this work we presented an audio-visual ASR system using deep RNNs trained with CTC objective function. We described a feature processing step for visual features using deep bottleneck layer and showed that it helps in faster convergence of RNN model during training. We presented a training protocol in which either of the modalities is turned off during training in order to avoid dependency on a single modality. Our results indicate that the trained model is robust to noise. In addition, we compared fusion strategies at the feature level and at the decision level. 

While the use of bottleneck features for visual modality helps in training, it requires frame level labels which involves an additional step of training audio RNN. Therefore, our system is not yet end-to-end. Our experiments in visual feature engineering with unsupervised methods like multi-modal auto-encoder\cite{ngiam2011multimodal} did not produce remarkable results. In future work we intend to explore other unsupervised methods for visual feature extraction such as canonical correlation analysis.

\bibliographystyle{splncs03}
\bibliography{references}

\end{document}